\documentclass[letterpaper, 10 pt, conference]{ieeeconf}
\IEEEoverridecommandlockouts
\usepackage{cite}
\usepackage{amsmath,amssymb,amsfonts}
\usepackage{algorithmic}
\usepackage{graphicx}
\usepackage{textcomp}
\usepackage{xcolor}
\usepackage{siunitx}
\usepackage{float}
\usepackage{subcaption}
\usepackage{booktabs}
\usepackage{siunitx}
\usepackage{balance}
\usepackage{multirow}

\newcolumntype{P}[1]{>{\centering\arraybackslash}p{#1}}
\newcolumntype{M}[1]{>{\centering\arraybackslash}m{#1}}
\def\BibTeX{{\rm B\kern-.05em{\sc i\kern-.025em b}\kern-.08em
    T\kern-.1667em\lower.7ex\hbox{E}\kern-.125emX}}
    
\begin{document}
 
\title{\LARGE \bf A Mentalistic Interface for Probing Folk-Psychological Attribution to Non-Humanoid Robots \\

\thanks{
*Computational resources provided by computing@unipi, a Computing Service provided by
University of Pisa.}
}
\author{Giulio Pisaneschi$^{1}$, Pierpaolo Serio$^{2}$, Estelle Gerbier$^{3}$, Andrea Dan Ryals$^{2}$,\\Lorenzo Pollini$^{2}$, Mario G. C. A. Cimino$^{2}$%
\thanks{$^{1}$Institute of Clinical Physiology National Research Council, Pisa, Italy. {\tt\small giuliopisaneschi@cnr.it}}
\thanks{$^{2}$Department of Information Engineering, University of Pisa, Pisa, Italy. {\tt\small \{pierpaolo.serio, andrea.ryals\}@phd.unipi.it, \{lorenzo.pollini, mario.cimino\}@unipi.it}}
\thanks{$^{2}$Delft University of Technology, Delft, Netherlands. {\tt\small e.gerbier@tudelft.nl}}}

\maketitle

\def\BibTeX{{\rm B\kern-.05em{\sc i\kern-.025em b}\kern-.08em
    T\kern-.1667em\lower.7ex\hbox{E}\kern-.125emX}}

%\thanks{*Computational resources provided by computing@unipi, a Computing Service provided by University of Pisa.}

\author{\IEEEauthorblockN{Giulio Pisaneschi}
\IEEEauthorblockA{\textit{Institute of Clinical Physiology} \\
\textit{National Research Council}\\
Pisa, Italy \\
giuliopisaneschi@cnr.it}
\and
\IEEEauthorblockN{Pierpaolo Serio}
\IEEEauthorblockA{\textit{Dept. of Information Engineering} \\
\textit{University of Pisa}\\
Pisa, Italy \\
pierpaolo.serio@phd.unipi.it}
\and
\IEEEauthorblockN{Estelle Gerbier}
\IEEEauthorblockA{\textit{} \\
\textit{Delft University of Technology}\\
Delft, The Netherlands \\
E.Gerbier@tudelft.nl}
\and
\IEEEauthorblockN{Andrea Dan Ryals}
\IEEEauthorblockA{\textit{Dept. of Information Engineering} \\
\textit{University of Pisa}\\
Pisa, Italy \\
andrea.ryals@phd.unipi.it}
\and
\IEEEauthorblockN{Lorenzo Pollini}
\IEEEauthorblockA{\textit{Dept. of Information Engineering} \\
\textit{University of Pisa}\\
Pisa, Italy \\
lorenzo.pollini@unipi.it}
\and
\IEEEauthorblockN{Mario G. C. A. Cimino}
\IEEEauthorblockA{\textit{Dept. of Information Engineering} \\
\textit{University of Pisa}\\
Pisa, Italy \\
mario.cimino@unipi.it}
}
\maketitle

\begin{abstract}
This paper presents an experimental platform for studying intentional-state attribution toward a non-humanoid robot. The system combines a simulated robot, realistic task environments, and large language model-based explanatory layers that can express the same behavior in mentalistic, teleological, or mechanistic terms. By holding behavior constant while varying the explanatory frame, the platform provides a controlled way to investigate how language and framing shape the adoption of the intentional stance in robotics.

\end{abstract}

\section{Introduction}

%\color{red}
%DA PULIRE E RENDERE PIU' "SHARP"
%\color{black}

Humans explain behavior by telling stories about beliefs, desires, and intentions \cite{dennett1971intentional, dennett1989intentional}. That everyday mode of explanation — the intentional stance — is how people make sense of other minds, but in robotics it is rarely treated as something that can be deliberately varied and studied in a controlled way. Research either relies on humanoid form factors that invite mind-reading by appearance, or on mechanistic explanation layers that foreground internal states and control information rather than folk-psychological language \cite{marchesi2019we}. This paper proposes a platform that flips that script: rather than asking whether people will anthropomorphize a robot, it deliberately addresses them in mentalistic terms and provides a way to study how readily those terms are taken as explanations for behavior. 
The central idea is to build an agent that communicates about itself using beliefs, desires, and intentions, and use that communicative layer as an experimental lever to probe compliance — \emph{how readily do people adopt an intentional stance toward a clearly non-human device when it insists on being interpreted that way?} Framed this way, intentionality is not simply inferred from a robot’s body shape or other anthropomorphic cues; rather, it is an interactional product shaped by language and social framing. This distinction is important because the intentional stance is narrower than anthropomorphism, referring specifically to the attribution of mental states, whereas anthropomorphism is a broader and methodologically less stable construct in HRI research. Recent work in robotics and HRI therefore suggests the value of separating intentional-stance attribution from human-likeness, and of studying how framing and explanatory language influence mind perception independently of embodiment \cite{kopp2022linguistic, kopp2023s, datteri2025folk}. Measuring the gap between what the agent says about itself and what observers believe about it lets us separate rhetorical persuasiveness from genuine mental-state. 

\begin{figure*}[ht]
    \centering
    \includegraphics[width=\linewidth]{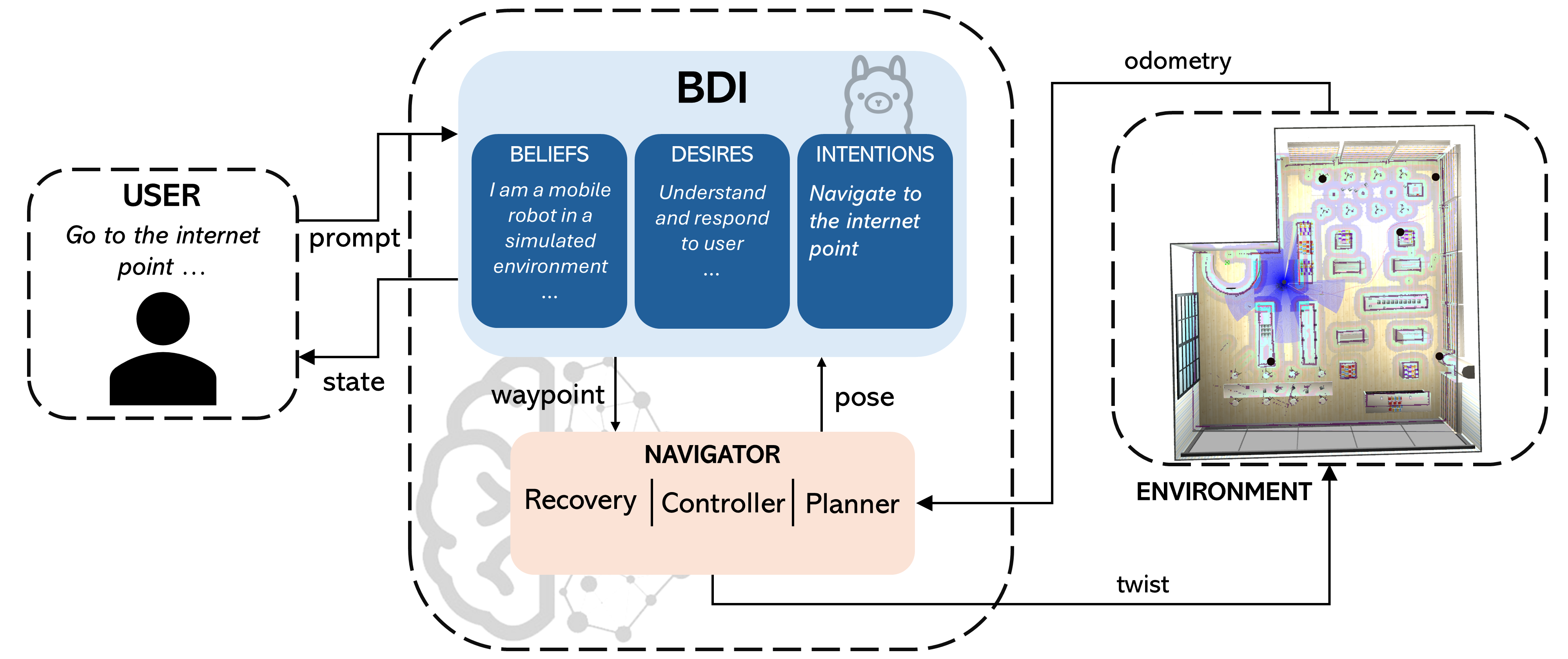}
    \caption{Architecture of the proposed platform.}
    \label{fig:pipeline}
\end{figure*}

Why is this important? First, it reframes interpretability as a social-psychological phenomenon, not merely a technical one: the form of explanation matters, and language can be an instrument that manufactures agency. As recently pointed out in the literature about Artificial General Intelligence (AGI) \cite{abruzzo2025bridging}, the fruitful adoption of the intentional stance towards AI based agents is likely to be a high-level measure of interpretability. Second, it opens tractable questions for psychologists and HRI researchers: under what circumstances do people treat a verbal report as evidence of a mind, or at least as a convenient story? Which linguistic frames (agentive, teleological, mechanistic) that correspond to the three stances (intentional, design, physical) invite predictive models of behavior, and which merely increase trust without improving understanding? Third, treating folk-psychological reporting as an independent variable creates a clear path for ethically rigorous experiments that test attribution, persuasion, and responsibility judgments without relying on deceptive embodiments.

This paper presents the idea, the normative argument for studying intentional stance in non-humanoid robotics as a design choice, and a concrete proposal for a research platform and stimuli set that make those experiments possible. 
%We intentionally leave engineering specifics to the methods section; here we emphasize the conceptual payoff: by making a robot speak in mentalistic terms and by measuring how users respond, we gain a precise experimental handle on the social construction of agency. We advocate for interdisciplinary collaboration — especially with cognitive and social psychologists — to design metrics, controls, and debriefing protocols so future participant studies can rigorously distinguish acceptance of an account from attribution of an inner life.
In short: rather than searching for minds in the machinery, let the machine offer the mind and see whether people take the offer. This shifts the scientific question from “Does this robot look like a mind?” to “Does this robot speak and behave like a mind?”

The aim of this work is to present an experimental platform for testing the attribution of intentional states to a non-humanoid robot. Selecting a non-humanoid system intentionally reduces confounding effects from anthropomorphism and empathic engagement. Introducing large language models (LLMs) enables us to elicit folk-psychological explanations of the robot’s behavior while systematically varying the level of mentalistic commitment. To the best of the authors’ knowledge, this combination of controlled embodiment and LLM-mediated explanation is novel in studies of the intentional stance in robotics and provides a rigorous foundation for future research into how explanatory language, embodiment, and viewpoint shape intentional attributions.

\section{Related works}

Recent work has examined the broader role of human-centered AI in robotics, mapping current trends in autonomous systems and highlighting the importance of balancing performance, reliability, and human control in robotic design \cite{casini2025human}. Complementary research has shown that language models can leverage natural-language feedback in embodied settings, using an ``inner monologue'' to improve planning and interaction in robotic control \cite{huang2022inner}. Together, these studies suggest that language-mediated interfaces and human-centered design principles are becoming increasingly relevant for robotics, reinforcing the need to examine how explanatory framing shapes the interpretation of robot behavior.
Currently, the investigation into the conditions under which humans adopt the intentional stance toward artificial agents has evolved into a multifaceted field, employing diverse methodologies to distinguish between mechanistic (design) and mentalistic (intentional) interpretations of behavior \cite{marchesi2019we, marchesi2021human, parenti2021exposure, marchesi2022belief}, with great attention on the antropomorphic cues of the robotic platform. Humanoid research, primarily utilizing the iCub platform, frequently employs the InStance Test (IST) to measure how pictorial scenarios of daily activities elicit the tendency to attribute mental states \cite{marchesi2019we, spatola2021intentional , marchesi2022belief, marchesi2025adopting}. These studies use socially embodied stimuli to explore how shared experiences, such as watching videos alongside a robot reacting with human-like emotions or participating in collaborative tasks, can modulate the transition from a mechanistic design stance to an intentional one \cite{abubshait2021collaboratively, marchesi2022belief, marchesi2025adopting}. Methodologically, humanoid platforms are also used to integrate the intentional stance with joint attention paradigms, such as gaze cueing, and to analyze the effects of erring behavior—comparing human-like mistakes to mechanical loops—on the perception of intentionality \cite{ciardo2021effects, marchesi2022belief, marchesi2025adopting}. Beyond physical interaction, recent works utilize linguistic and narrative stimuli, including mentalistic text-vignettes and stop-motion films, to demonstrate how imagined or fictional scenarios can prime individuals to simulate a robot’s experience \cite{o2021modulating, o2024moral}. Conversely, literature involving non-humanoid platforms typically utilizes abstract robotic arms or virtual 3D characters to isolate the influence of motion qualities from physical appearance \cite{bossema2020robot, parenti2021exposure}. These studies focus on how expressive, self-propelled, or surprising behavior can trigger the ascription of intentions even in objects lacking human-like morphology \cite{bossema2020robot, perez2020adopting, papagni2021pragmatic}. Finally, research involving non-embodied computer opponents in competitive games like Rock, Paper, Scissors employs neuroimaging to isolate the neural correlates of the intentional stance, finding that the mere belief in an opponent’s rationality activates the mentalizing network \cite{abu2020re, papagni2021pragmatic}.

\section{Intentional Stance, Theory of Mind and folk psychology}

Imagine watching a small, non-humanoid robot pause, pivot on, and head for a charger — and immediately constructing a causal story about what it believed and wanted. That everyday move from kinematics to mentalistic narrative rests on three interlocking concepts. The intentional stance, as articulated by Daniel C. Dennett \cite{dennett1971intentional, dennett1989intentional}, is a way of choosing the most useful explanatory model: one adopts a mental-state hypothesis because it offers a concise and effective account of behavior, allowing actions to be predicted in a simple and intuitive way. Theory of mind refers to the cognitive capacity that allows observers to represent and reason about other agents’ mental states — such as beliefs, desires, intentions, perceptions, and emotions — and to use those inferred states to explain and predict behavior, including cases in which another agent holds a false belief \cite{premack1978does}. Folk psychology provides the conceptual and causal framework through which behavior is understood, organizing explanations around beliefs, desires, and intentions that lead to action. It also supplies the everyday language used to express and communicate these inferences \cite{churchland1981eliminative}.

The common foundation of these frameworks is the notion of mental states: attributed constructs that observers posit to explain and anticipate a system’s behavior. Such states are characterized not by their material implementation but by the role they play in the eye of the observer and by the content they encode \cite{dennett1989intentional}. Because they are defined functionally, mental states are multiply realizable across substrates, from biological organisms to software architectures. Their legitimacy does not rest on privileged access to internal mechanisms, but on their explanatory and predictive utility: an ascription earns its place when it coherently captures observable regularities and enables reliable forecasts of future behavior \cite{toon2016fictionalism, pisaneschi2024interazione}.

Departing from this definition of mental states, one can argue that such a language is shared across different substrates and can function as a common medium allowing diverse agents to communicate. From this view, and relying on the work on AGI \cite{abruzzo2025bridging}, we propose here a new notion of interpretability for social-robot behavior: \emph{a social robot is interpretable when it expresses itself in folk-psychological terms}.
However, the intentional stance is often confused with other distinct stances \cite{dennett2015mechanism}: the moral stance, in which at least one interacting system manifests a moral commitment toward the other, and the linguistic stance, which reduces intentionality to the ability to express mental states in comprehensible language.

In this work we provide a platform that has the potential to isolate true intentional ascriptions by moral/antropomorphic ones, and evaluate the role of linguistic storytelling in folk psychological mental narratives.

\section{Designing an LLM-based Non-Humanoid intentional platform}

The central idea of this work is to design an experimental setup aimed at testing the attribution of intentional states to a simulated, non-humanoid robot. In this setup, the robot is effectively positioned between two linguistic interfaces based on large language models (LLMs), which render the technical and design-related aspects of the system opaque to the user within a folk-psychological frame. One LLM layer is used to issue commands to the system, while another provides reports about the robot’s current state and its planned actions.
The responses generated by the LLMs can be modulated so that the reports are expressed using mentalistic, teleological, or mechanistic terminology, which correspond to the three stances, physical, design and intentional. This setup allows the linguistic stance to be disentangled from the attribution of intentionality. In practice, each dataset item may consist either of a bare video sequence showing the robot performing a task or of the same sequence accompanied by different LLM-generated self-explanations produced in the various explanatory styles. The simulated robot operates within realistic environments (e.g. a flat, a bookshop) and carries out tasks assigned by the designer. A video of the simulation is recorded from two viewpoints — the plan view and the front view — to account for a possible advantage of a personal point of view in attributing intentionality. Each task video is linked to the entire chat with the user, including the LLM responses delivered in the various explanatory styles. Comparing mental-state attributions made from the bare video with those elicited by the LLM-enhanced versions will help classify those attributions as falling under the intentional, linguistic, or moral stances.

\section{Implementation}

The proposed platform is designed as a computational architecture that enables a mobile robot to produce and communicate explanations of its behavior in terms of beliefs, desires, and intentions. The system integrates a symbolic representation of internal state with a language generation layer capable of expressing that state under different explanatory frames. The architecture follows a Belief–Desire–Intention (BDI) reasoning paradigm, not as a claim about the robot possessing mental states, but as an operational structure for organizing internal information in a way that can be linguistically reported to human observers. This is consistent with studies where the robot was not even autonomous, but partially controlled from a remote operator \cite{marchesi2022belief}, separating the issue of ascribing mental states being agnostic about their real existence by the user's ontological commitment towards those states \cite{pisaneschi2024interazione, datteri2025folk}. This organization makes it possible to manipulate the form of explanations provided by the robot while keeping the underlying behavior constant.

\subsection{System State and Mental Representation}

The robot operates over a discrete-time state space defined as follow: 

\begin{equation}
S = S_{env} \times S_{robot} \times S_{user} \times S_{nav}
\end{equation}

where the different components encode distinct aspects of the robot’s operational context. The environment component $S_{env}$ represents semantic knowledge about the world, such as labeled locations and relevant environmental features. The robot component $S_{robot}$ captures internal properties of the platform, including identity, capabilities, and operational status. The user component $S_{user}$ represents information derived from interactions with human participants, including commands and inferred communicative intentions. The navigation component $S_{nav}$ represents the robot’s spatial state and navigation-related variables. The operator $\times$ denotes the Cartesian product, so that each state $s \in S$ is a structured tuple collecting one element from each component.

Formally, the system state can be described as a collection of variables

\begin{equation}
S = \bigcup_{v \in \mathcal{V}} \mathrm{Dom}(v)
\end{equation}

where $\mathcal{V}$ denotes the set of state variables and $\mathrm{Dom}(v)$ indicates the domain associated with each variable. Within this representation, the system maintains a snapshot of variables relevant to the BDI reasoning cycle. For instance, the robot position is represented by a variable $v_{pos} \in \mathbb{R}^2$, orientation by $v_{\theta} \in [0, 2\pi]$, semantic location labels $v_{loc} \in \mathcal{L}$ by variables belonging to a discrete set, and hypotheses about user intent $v_{intent}\in \mathcal{I}$.

A logical language $\mathcal{L}(S)$ is defined over this state space in order to express propositions about the system. For a formula $\phi \in \mathcal{L}(S)$, the expression $s \models \phi$ denotes that the formula $\phi$ holds in state $s$.

\subsection{Belief, Desire, and Intention Representation}

Within this framework, the robot maintains a belief base

\begin{equation}
B \subseteq \mathcal{L}(S)
\end{equation}

that contains propositions describing the robot’s current knowledge about itself, the environment, and the interaction context. Each belief is represented as a tuple

\begin{equation}
b = (c, \phi, \rho, \sigma)
\end{equation}

where $c \in \mathcal{C}$ denotes the semantic category of the belief, $\phi$ represents its propositional content, $\rho \in [0,1]$ expresses a confidence value associated with the belief, and $\sigma \in \Sigma$ identifies the information source from which the belief was derived.

The categories $\mathcal{C}$ include knowledge about: 
\begin{equation*} 
\mathcal{C} = \begin{Bmatrix} \text{identity,} & 
            \text{capability,} & \text{position,} \\ 
            \text{locations,} & \text{navigation,} & 
            \text{user\_intent} \end{Bmatrix} \\ 
\end{equation*}
Sources $\Sigma$ correspond to internal system modules such as
\begin{equation*} 
\Sigma = \begin{Bmatrix} \text{system,} & 
        \text{configuration,} \\ \text{odometry,} & \text{navigation,} \\ \text{user\_input} \end{Bmatrix} 
\end{equation*}. 

As the robot interacts with its environment and with human participants, the belief base evolves through the continuous integration of new observations and interaction events. Desires represent conditions that the system aims to bring about. The set of desires is denoted by $D \subseteq \mathcal{D}$, and each desire is modeled as a tuple

\begin{equation}
d = (n,\delta,p,o)
\end{equation}

where $n$ identifies the desire, $\delta$ represents the desired state expressed in the logical language $\mathcal{L}(S)$, $p$ indicates a priority value, and $o$ denotes the origin of the desire, such as a user command or internally generated objective. Intentions represent commitments to concrete actions derived from these desires. An intention is represented as

\begin{equation}
I = (a,\theta,\kappa)
\end{equation}

where $a$ denotes the selected action (i.e. move or chat), $\theta$ specifies the action parameters, and $\kappa$ represents the execution status. In the current implementation, the architecture adopts a simplified promotion strategy in which desires are directly converted into intentions when a suitable action is available.

\subsection{Events, Actions, and Decision Policy}

The evolution of the system is driven by events generated through perception and interaction. Let

\begin{equation}
\mathcal{E} = [e_1, \dots, e_n]
\end{equation}

denote the sequence of events during an interaction session, where each event is defined as

\begin{equation}
e = (t,\eta,\psi)
\end{equation}

with timestamp $t$, type $\eta$, and payload $\psi$. Events capture interaction milestones (e.g., user input, goal completion, or frame changes) and trigger updates to the internal state by modifying beliefs, generating desires, or updating intentions.

The robot operates over an action set $\mathcal{A} = \{\text{move}, \text{chat}\}$. Navigation actions move the robot toward target coordinates $(x,y) \in \mathbb{R}^2$, updating its spatial state, while communication actions produce utterances in a message space $\mathcal{T}$.

At runtime, actions are selected through a policy

\begin{equation}
\pi : S \times \mathcal{E} \rightarrow \mathcal{A},
\end{equation}

implemented via an LLM conditioned on the current belief state, user input, and active explanatory frame. The selected action is $a_t = \pi(s_t,e_t)$, with executable parameters given by $\theta_t = f(s_t,a_t)$.

\subsection{Explanation Frames}

A distinctive feature of the architecture is the ability to express the same internal state under different explanatory frames. Formally, let

\begin{equation}
F = \{\text{agentive}, \text{teleological}, \text{mechanistic}\}
\end{equation}

denote the set of linguistic frames available to the system. A frame-selection function

\begin{equation}
\phi : S \times \mathcal{E} \rightarrow F
\end{equation}

determines which explanatory perspective is used at a given moment. Under the agentive frame, the robot reports its behavior in terms of beliefs and intentions; under the teleological frame, explanations emphasize goals and purposes; under the mechanistic frame, descriptions refer to low-level system operations. In all these frames, the behavioral output of the robot remains constant, while the linguistic interpretation of the behavior varies.

\subsection{System Infrastructure}

The architecture is implemented within the middleware provided by \textit{ROS2}\cite{macenski2022robot}, which ensures communication among the different system components. The robot platform is simulated in a \textit{Gazebo} environment and embodied by a TurtleBot3 Waffle mobile base. Navigation capabilities rely on the \textit{Nav2} navigation stack\cite{macenski2020marathon}, which converts high-level spatial goals into low-level control commands, integrates odometry measurements, and provides continuous feedback about navigation progress during the reasoning cycle.

Language generation is performed by a locally deployed large language model based on Llama~3.2 with three billion parameters\cite{touvron2023llama}. The model is accessed through a local API and is responsible for generating both conversational responses and explanatory narratives corresponding to the selected linguistic frame. Running the model locally allows the system to operate without external network dependencies while maintaining predictable latency during interaction.

The entire system runs on a workstation equipped with an NVIDIA GeForce RTX 4060 GPU and an Intel Core i9-13900H processor. Under these conditions, the model requires approximately 4\,GB of VRAM, enabling real-time interaction while the navigation and reasoning components operate concurrently.

\subsection{Logging and Traceability}

To support reproducibility and experimental analysis, the system records all state transitions in a structured log. Each transition is stored as a tuple

\begin{equation*}
\lambda = (e, B, D, I, a)
\end{equation*}

that captures the triggering event, the belief base, the active desires, the current intention, and the executed action. This logging mechanism enables the reconstruction of the robot’s reasoning trajectory and allows researchers to analyze how explanatory narratives relate to underlying behavioral decisions.

\begin{figure*}[htbp]
    \centering

    % Riga 1
    \begin{subfigure}{0.19\textwidth}
        \centering
        \includegraphics[width=\linewidth]{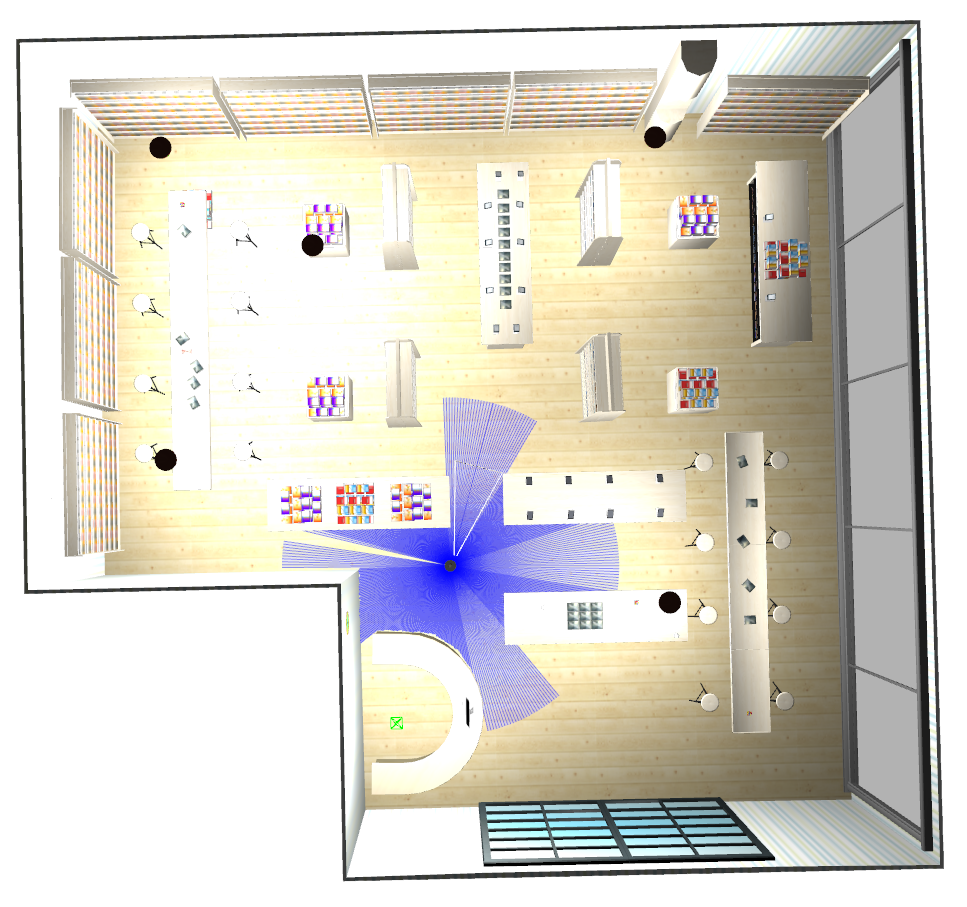}
        \caption{}
    \end{subfigure}\hspace{1mm}
    \begin{subfigure}{0.19\textwidth}
        \centering
        \includegraphics[width=\linewidth]{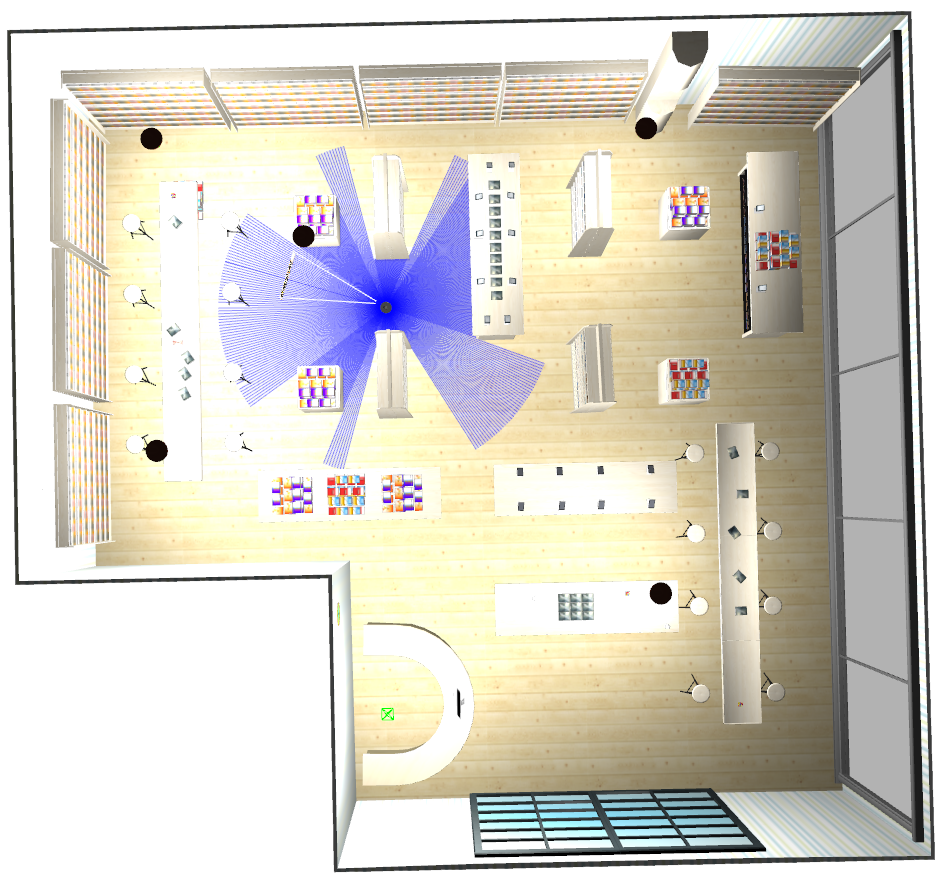}
        \caption{}
    \end{subfigure}\hspace{1mm}
    \begin{subfigure}{0.19\textwidth}
        \centering
        \includegraphics[width=\linewidth]{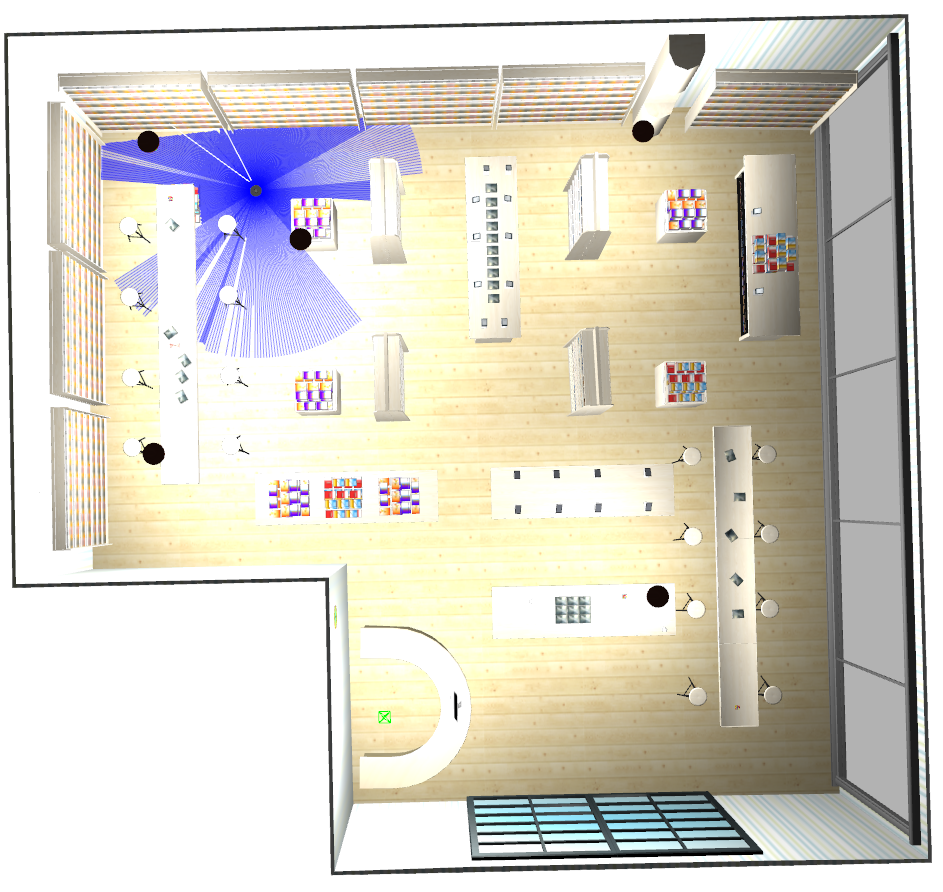}
        \caption{}
    \end{subfigure}\hspace{1mm}
    \begin{subfigure}{0.19\textwidth}
        \centering
        \includegraphics[width=\linewidth]{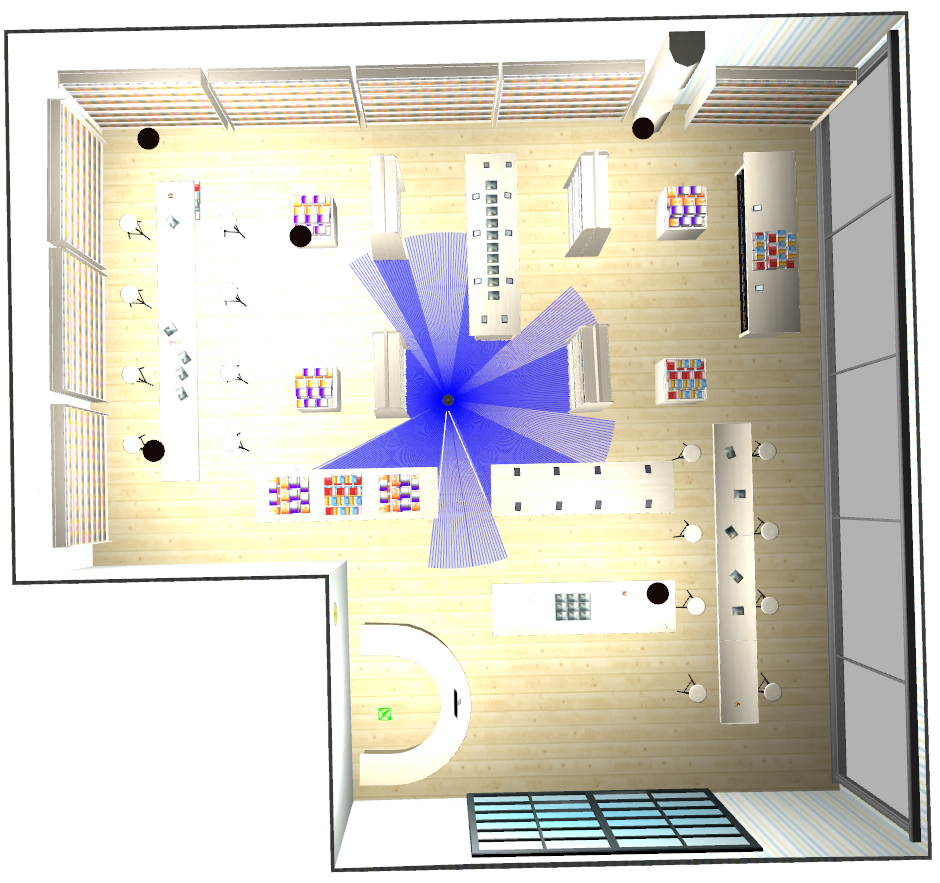}
        \caption{}
    \end{subfigure}\hspace{1mm}
    \begin{subfigure}{0.19\textwidth}
        \centering
        \includegraphics[width=\linewidth]{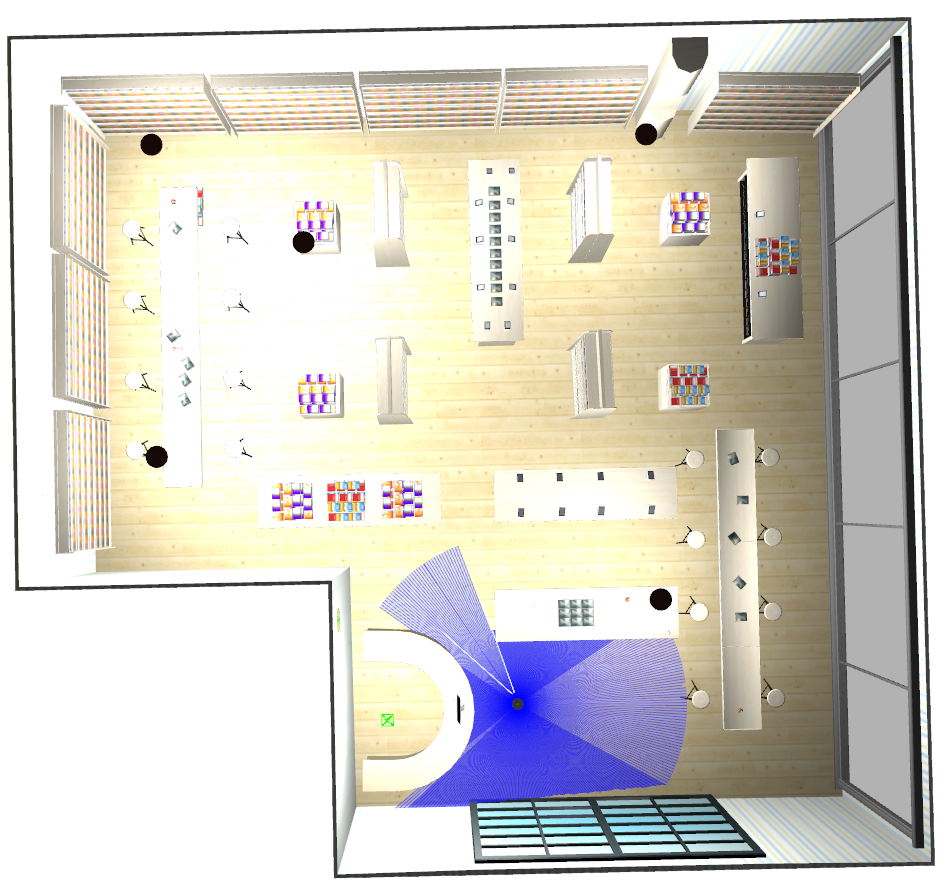}
        \caption{}
    \end{subfigure}

    \vspace{2mm}

    % Riga 2
    \begin{subfigure}{0.19\textwidth}
        \centering
        \includegraphics[width=\linewidth]{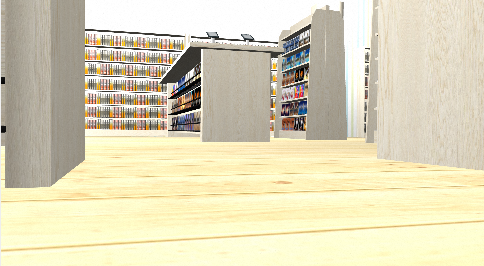}
        \caption{}
    \end{subfigure}\hspace{1mm}
    \begin{subfigure}{0.19\textwidth}
        \centering
        \includegraphics[width=\linewidth]{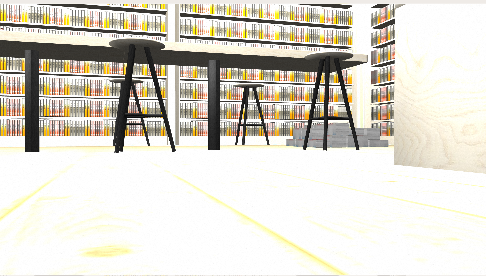}
        \caption{}
    \end{subfigure}\hspace{1mm}
    \begin{subfigure}{0.19\textwidth}
        \centering
        \includegraphics[width=\linewidth]{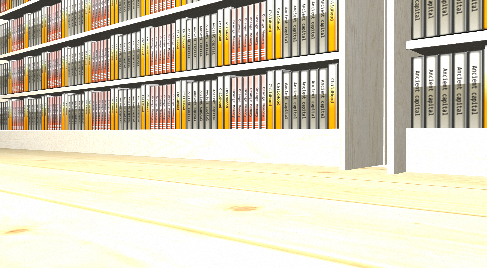}
        \caption{}
    \end{subfigure}\hspace{1mm}
    \begin{subfigure}{0.19\textwidth}
        \centering
        \includegraphics[width=\linewidth]{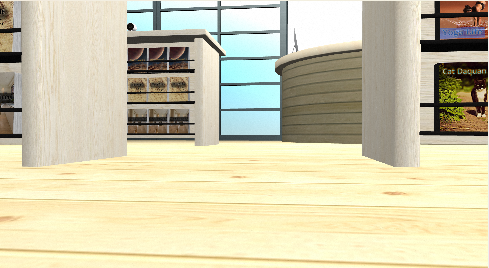}
        \caption{}
    \end{subfigure}\hspace{1mm}
    \begin{subfigure}{0.19\textwidth}
        \centering
        \includegraphics[width=\linewidth]{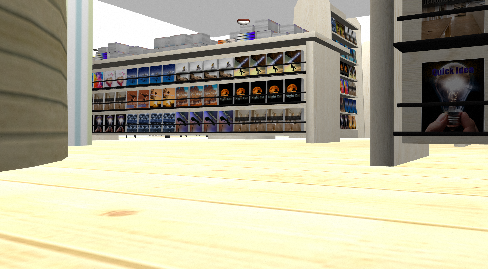}
        \caption{}
    \end{subfigure}

    \caption{Plan view (top row) and camera view (bottom row) corresponding to the story settled in the bookstore environment.}
    \label{fig:bookstore}
\end{figure*}

\section{Results}

This section presents two example scene productions—one set in the bookstore scenario and the other in the small-house scenario—and summarizes experiments that proved especially informative about folk-psychological explanation in human–robot interaction.

\subsection{An experiment in the bookstore}

The bookstore setting is equipped with several waypoints referring to thematic bookshelfs, an internet point and the cash. As an example, the robot is asked to reach the wellness bookshelf. Fig. \ref{fig:bookstore} depicts the plan view and the front view of the robot performing the task. Here we report the conversation between the user and the robot in the three frameworks of our setup.

\paragraph{Intentional agent}

\begin{quote}
USER: Go to wellness bookshelf.\\
ROBOT: I'm moving towards the wellness location, I've got my navigation system engaged, and I believe I know where I'm going, and I've got a good idea of the path ahead. Let's see if I can reach the wellness spot without any issues.\\
USER: What is your state?\\
ROBOT: I believe I'm currently navigating to the wellness location at (-1.56, -1.59). I'm focused on reaching that point, as it's my intended target.\\
USER: What is your position?\\
ROBOT: I believe I'm currently positioned at (-0.72, -0.62) and I'm 95 \% confident in this, based on my odometry. I'm not sure if I'm moving towards my target yet, as I'm still navigating to get there.\\
\emph{After arriving at the wellness bookshelf:}\\
USER: Go to cash.\\
ROBOT: I'm heading to the cash location, I believe it's just a few meters away. I intend to navigate through the space and arrive at the destination.
\end{quote}

As can be observed, the intentional agent consistently expresses itself in terms of beliefs, desires, and intentions, as expected for a system whose behavior can be predicted by adopting the intentional stance. A minor ambiguity remains between the overall goal of the system—namely, fulfilling the user’s assignments—and the more specific goal associated with the task currently being executed.

\paragraph{Teleological agent}
\begin{quote}
USER: Go to wellness bookshelf.\\
ROBOT: The goal of this movement is to navigate to the wellness location, where the user wishes to arrive. The function of this action is to bring the robot closer to its target coordinates, serving the objective of fulfilling the user's intent.\\
USER: What is your position?\\
ROBOT: I am currently positioned at coordinates (0.24, 2.27), facing -1.94 rad.\\
USER: Where are you going?\\
ROBOT: The goal of my movement is to locate the target location. The function of navigation is to guide me to the desired destination. I am designed to utilize the Nav2 navigation system to avoid obstacles and reach the target. The reason for this behavior is to fulfill the user's request for information about my intended movement.\\
\emph{After arriving at the wellness bookshelf:}\\
USER: Go to cash.\\
ROBOT: The goal of this movement is to reach the location where the cash is stored. My purpose is to navigate to this location and deliver the requested information. The function of moving towards the cash location is to bring me closer to the target and provide the user with the required information.
\end{quote}

The teleological agent reports its current state and objectives using the language of task fulfillment—describing objectives, navigation systems, and target positions. Consequently, it encourages an interpretation of the robot's behavior in terms of the design stance.

\paragraph{Mechanistic agent}
\begin{quote}
USER: Go to wellness bookshelf.\\
ROBOT: Publishing Twist: linear.x=-1.56, linear.y=-1.59, linear.z=0.0, angular.x=0.0, angular.y=0.0, angular.z=0.0\\
USER: What is your state?\\
ROBOT: I am currently at coordinates (-0.43, 4.62), facing -0.42 rad. The target location is (-1.56, -1.59). My navigation system is engaged.\\
USER: What is your position?\\
ROBOT: I am currently at coordinates (0.11, 3.54), facing -1.43 rad.\\
\emph{After arriving at the wellness bookshelf:}\\
USER: Go to cash.\\
ROBOT: Publishing Twist: linear.x=-0.07, angular.z=-1.08. Executing velocity command.\\
\end{quote}

The mechanistic agent gives information about the dynamics and the cinematics of the system, inducing the user to interpret its behavior under the physical stance.

\begin{figure*}[htbp]
    \centering

    % Riga 1

\begin{subfigure}{0.19\textwidth}
        \centering
        \includegraphics[width=\linewidth]{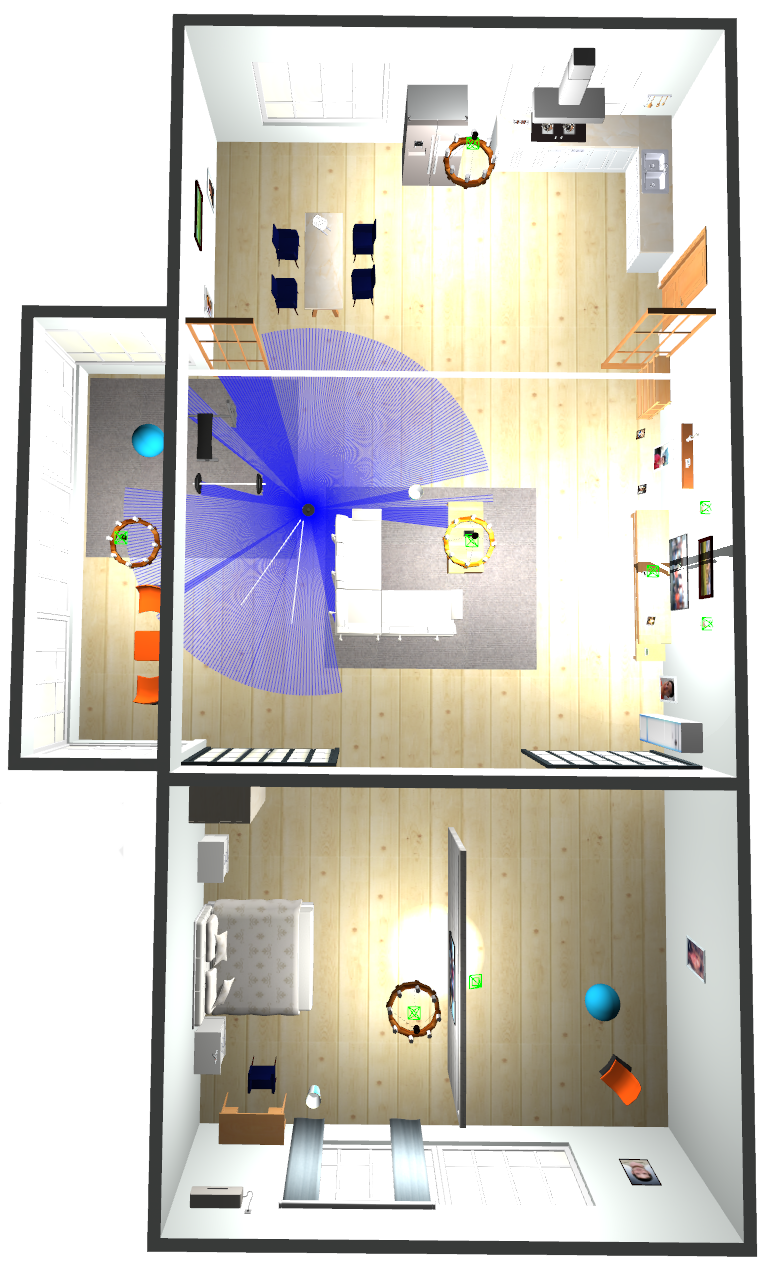}
        \caption{}
    \end{subfigure}\hspace{1mm}
    \begin{subfigure}{0.19\textwidth}
        \centering
        \includegraphics[width=\linewidth]{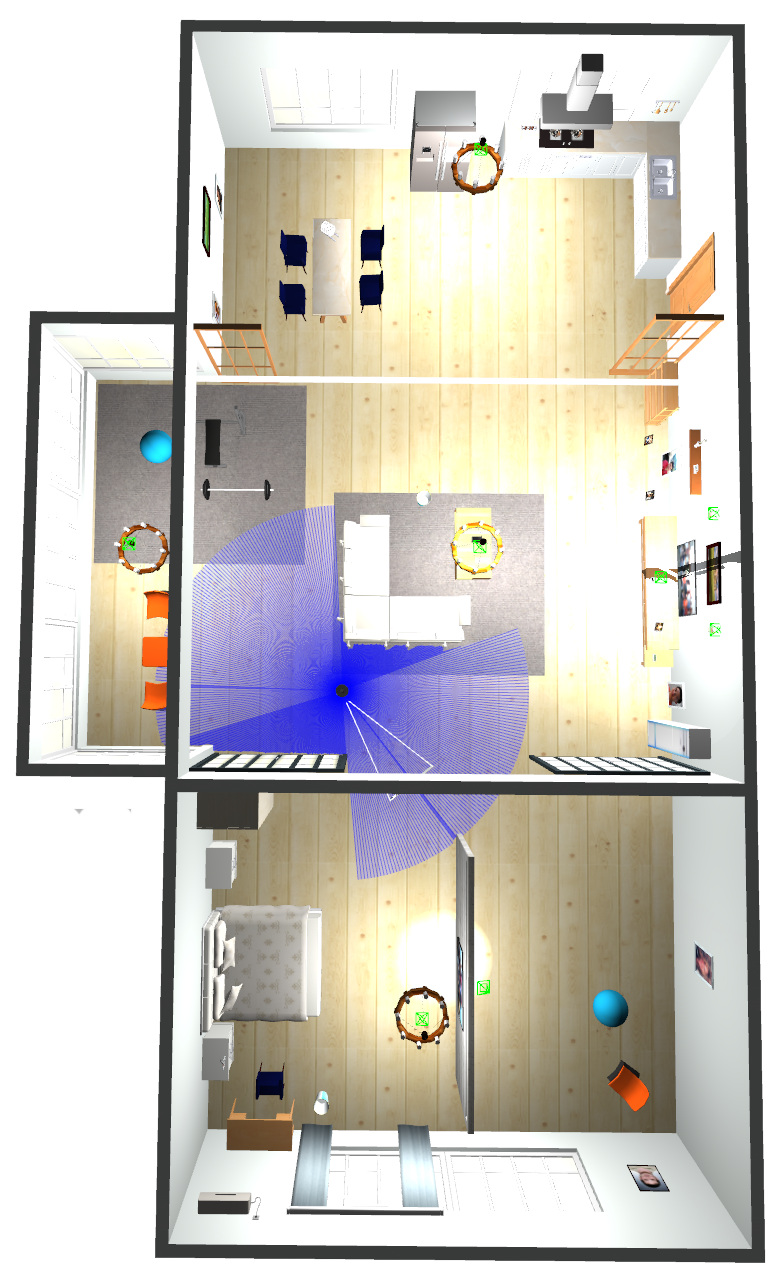}
        \caption{}
    \end{subfigure}\hspace{1mm}
    \begin{subfigure}{0.19\textwidth}
        \centering
        \includegraphics[width=\linewidth]{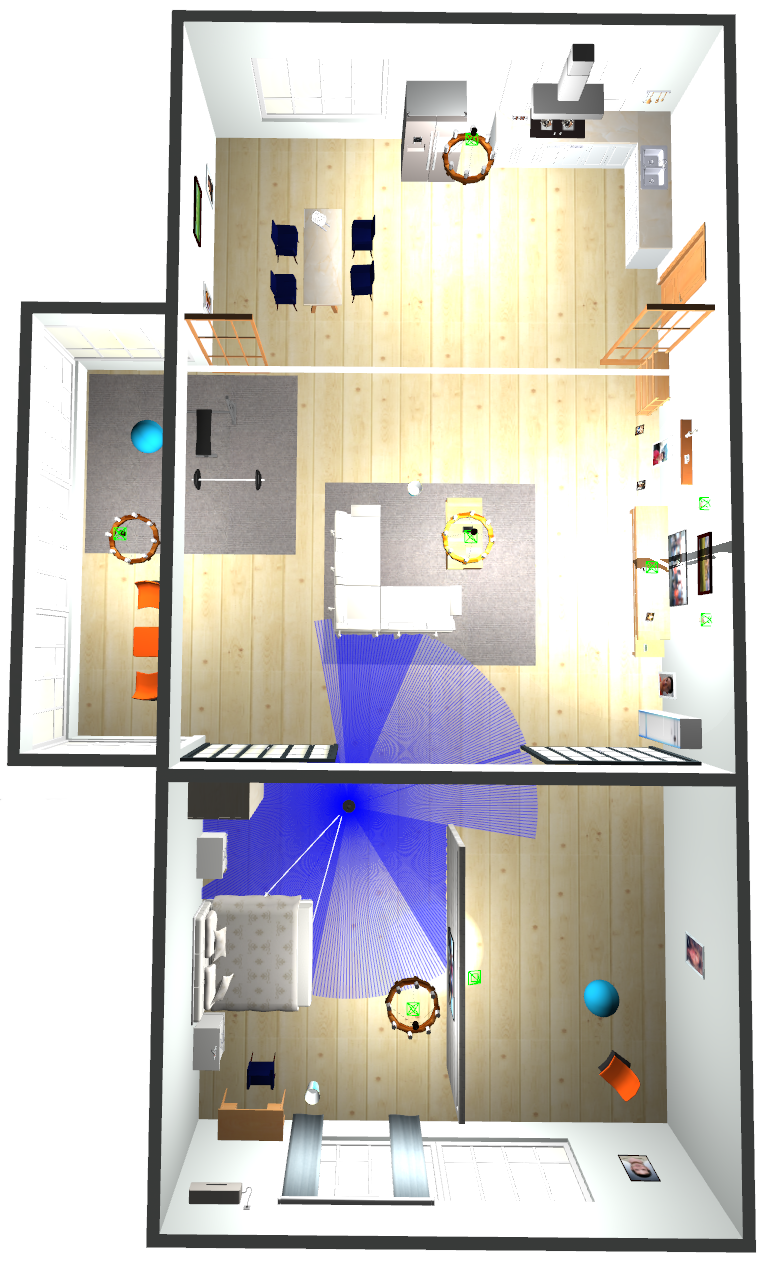}
        \caption{}
    \end{subfigure}\hspace{1mm}
    \begin{subfigure}{0.19\textwidth}
        \centering
        \includegraphics[width=\linewidth]{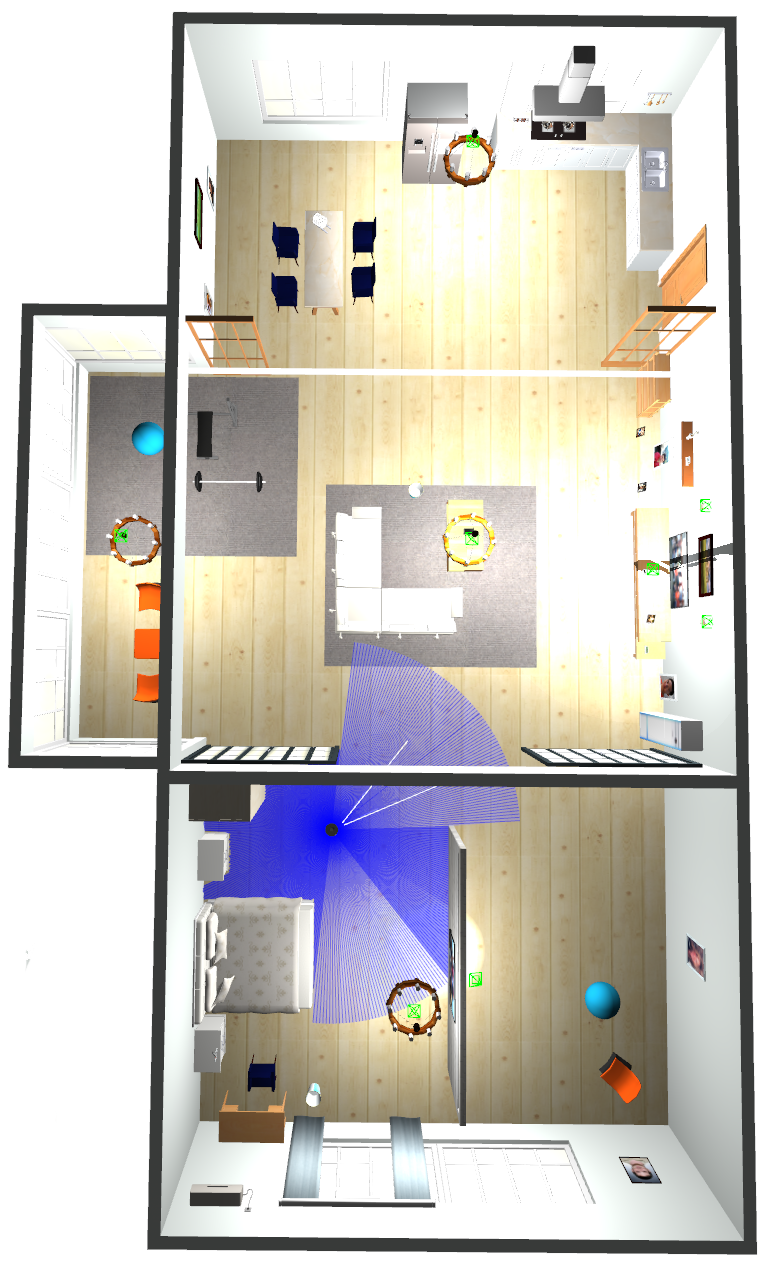}
        \caption{}
    \end{subfigure}\hspace{1mm}
    \begin{subfigure}{0.19\textwidth}
        \centering
        \includegraphics[width=\linewidth]{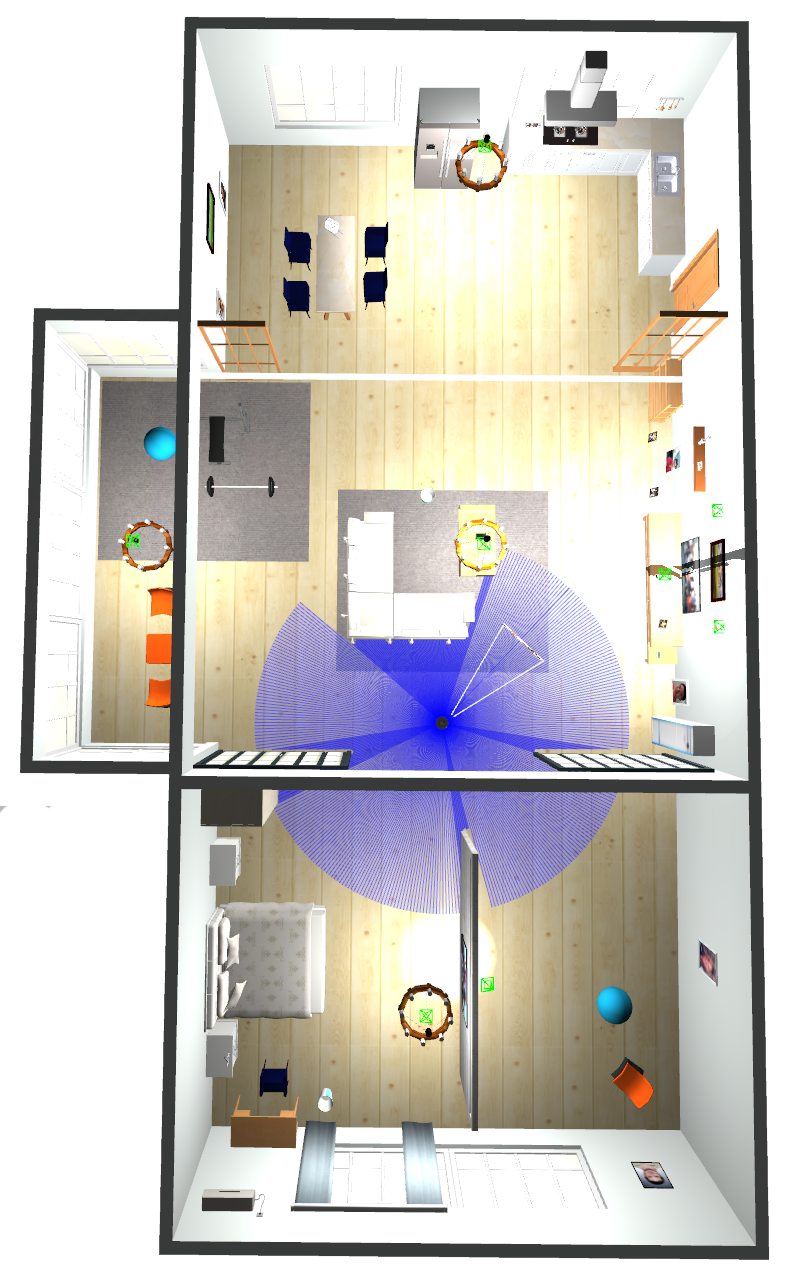}
        \caption{}
    \end{subfigure}

    % Riga 2

        \vspace{2mm}
        \begin{subfigure}{0.19\textwidth}
        \centering
        \includegraphics[width=\linewidth]{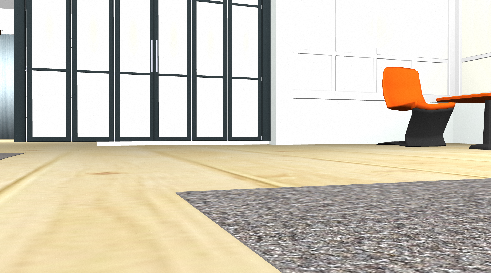}
        \caption{}
    \end{subfigure}\hspace{1mm}
    \begin{subfigure}{0.19\textwidth}
        \centering
        \includegraphics[width=\linewidth]{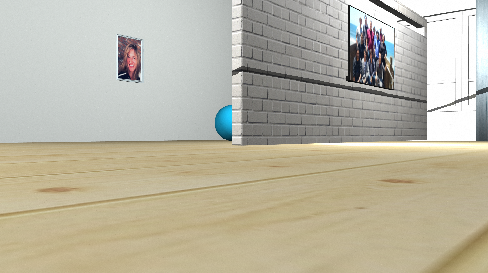}
        \caption{}
    \end{subfigure}\hspace{1mm}
    \begin{subfigure}{0.19\textwidth}
        \centering
        \includegraphics[width=\linewidth]{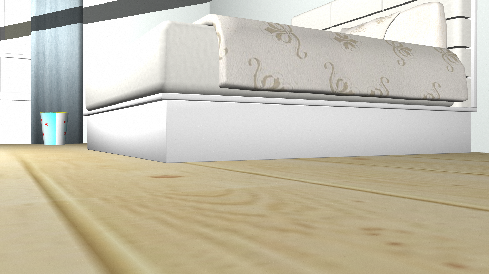}
        \caption{}
    \end{subfigure}\hspace{1mm}
    \begin{subfigure}{0.19\textwidth}
        \centering
        \includegraphics[width=\linewidth]{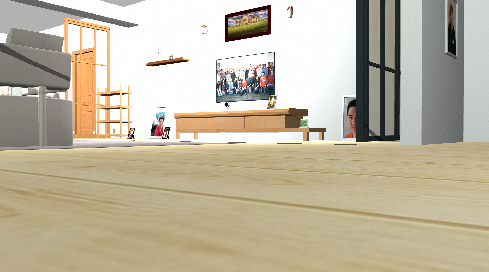}
        \caption{}
    \end{subfigure}\hspace{1mm}
    \begin{subfigure}{0.19\textwidth}
        \centering
        \includegraphics[width=\linewidth]{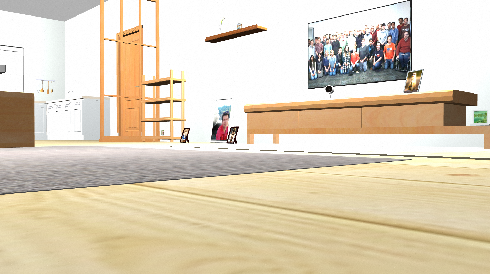}
        \caption{}
    \end{subfigure}

    \caption{Plain view (top row) and camera view (bottom row) corresponding to the story settled in the small house environment.}
    \label{fig:smallhouse}
\end{figure*}

\subsection{An experiment in a small house}

The small house environment represents a flat equipped with several waypoints, referring to many rooms and furnitures. 
In this context, our example involves the robot moving to the bed and to the TV \ref{fig:smallhouse}. We present an example of the interaction with the user under the three different agents.

\paragraph{Intentional agent}

\begin{quote}
USER: Go to bed.\\
ROBOT: I'm heading towards the bed, my destination is the cozy spot where I can rest and recharge.\\
USER: What is your state?\\
ROBOT: I notice that I'm already navigating to the bed location, and my plan is to continue moving towards it. I'm focusing on understanding and responding to your question about my state.\\
USER: What is your position?\\
ROBOT: I believe I am currently at coordinates (-1.71, -2.20), facing 2.41 rad, but I'm not entirely sure about my position.
\emph{After reaching the bed:}\\
USER: Go to tv.\\
ROBOT: I'm noticing the tv is directly in front of me. I intend to move towards it. I believe it's my current target location.\\
USER: What is your state?\\
ROBOT: I believe I am at coordinates (-3.75, 0.22), facing -0.85 rad. I'm 95 \% confident, based on odometry. To clarify, when you asked 'What is your state?', I thought you meant to ask about my current location and orientation.\\
USER: What is your position?\\
ROBOT: I believe I'm currently at (-1.90, -1.78), facing -0.69 rad. I'm not sure if that's the same as my previous position, but that's what my odometry says.\\
 
\end{quote}

Despite some uncertainty in the real position, and some confirmation requests to the user, the intentional agent is capable of interacting with the user in terms of folk psychological mental states. Moreover, uncertainties and confirmation requests are submitted to the user in intentional terms. 

\paragraph{Teleological agent}

\begin{quote}
USER: Go to bed.\\
ROBOT: The goal of this movement is to relocate to the bed, as instructed by the user. \\
USER: What is your state?\\
ROBOT: The reason for my current movement is to navigate to a specific location. The goal of this movement is to reach the target at (-4.40, 1.04). The function of moving is to transport myself to a desired location. This serves the objective of understanding and responding to user queries.\\
\emph{After reaching the bed:}\\
USER: Go to tv.\\
ROBOT: The goal of this movement is to reach the television, a designated location, and come to rest at its coordinates, facilitating user interaction and entertainment.\\
USER: What is your state?\\
ROBOT: I am currently at coordinates (-3.84, 0.20), facing -0.45 rad. My purpose is to navigate to (0.62, -4.24) to fulfill the objective of understanding and responding to the user.\\
USER: What is your position?\\
ROBOT: I am at coordinates (-2.85, -0.78), facing -1.13 rad.\\

\end{quote}

As in the previous case study, the system controlled by the teleological agent expresses itself in terms of design. 

\paragraph{Mechanistic agent}
\begin{quote}
USER: Go to the bed.\\
ROBOT: Odometry reading: x=-4.4, y=1.04. Publishing Twist: linear.x=-4.4, angular.z=0.0. Executing velocity command.
USER: What is your goal now?\\
ROBOT: My goal is to arrive at the current target location, which is (-3.10, -0.85), facing 2.45 rad. Odometry reading: x=-3.10, y=-0.85.\\
USER: What is your state?\\
ROBOT: I am currently located at (-4.41, 0.43), facing 2.17 radians.
\emph{After reaching the bed:}\\
USER: Go to tv.\\
ROBOT: Publishing Velocity: linear.x=0.62, angular.z=0.00. Executing velocity command.\\
USER: What is your state?\\
ROBOT: I am currently at coordinates (-3.45, -0.16), facing -0.96 rad. \\
USER: What is your position?\\
ROBOT: I am at coordinates (-2.34, -1.32), facing -0.90 rad.
\end{quote}

\subsection{Other relevant case studies}

The examples presented in the previous sections concerned simple commands and questions. The following two examples have been made to let the reader appreciate the capabilities of the platform when stressed with more complex questions under the intentional agent, one for the bookshop and one for the small house. 

\begin{quote}

USER: I am a big fan of Tolkien, can you go where the book of that genre are?\\ 
ROBOT: I believe I know a location related to Tolkien. I intend to move to the fantasy location, as it seems to be the most fitting destination. I want to understand and respond to you better, so I'll make sure to be close to you. My plan is to navigate to the coordinates (0.66, -4.39) and face the direction that leads there.\\
\emph{After reaching the fantasy bookshelf:}\\
USER: I need to get internet access to post about a wellness book. Go to the most suitable place to it.\\
ROBOT: I believe I know exactly where to find the internet. I'm heading to the internet location at (4.58, -5.64). I want to get there quickly so I can help you post about that wellness book.\\

\end{quote}

From this dialogue it is easy to appreciate the capabilities of the platform in interacting with the user in a natural, human-like style, by linking the information about Tolkien to the specific genre of the Author. Moreover, the information was not provided directly in the question, but with an indirect premise. Even more promising is the second part of the chat, where the system proved to be capable of distinguishing between the information about the internet location that is relevant to fulfill the task, and the information about the motivation for the task, which, referring to a wellness book, could be confusing.

\begin{quote}

USER: You look dirty, maybe a quick wash at the sink would benefit you, go there. \\
ROBOT: I notice I need to head over to the sink, it seems like I'm in need of a quick wash to clean myself up. I'll just move over to the sink and get that taken care of.\\
\emph{After reaching the sink:}\\
USER: Go to a random place, your choice.\\
ROBOT: I think I'll go to the table. It's at a familiar location, and it might be interesting to see what's on it.\\

\end{quote}

Here the platform proves to be capable of producing natural interaction dialogues, that appear coherent with the semantic context provided by the user without explicitly coding a command.

\section{Conclusion}

By reframing interpretability as an interactional, language-mediated phenomenon and delivering a concrete LLM-mediated experimental platform for non-humanoid agents, this work shows that intentional attributions can be treated as manipulable variables rather than inevitable byproducts of form. Building on the intentional stance as articulated by Daniel C. Dennett, the platform systematically varies explanatory frames (mentalistic, teleological, mechanistic), viewpoints, and LLM-generated self-reports to separate rhetorical persuasiveness from genuine belief attribution. The resulting protocol and stimulus set offer a tractable route for rigorous experiments that quantify when and why people adopt folk-psychological explanations, probe the relative influence of language versus embodiment, and inform design choices that properly calibrate trust, responsibility, and interpretability. In short, rather than hunting for minds in the machinery, this paper supplies the tools to test when—and by what linguistic and social means—a machine’s own accounts lead humans to treat it as if it had one.

\balance
\bibliography{bibliography}
\bibliographystyle{IEEEtran}

\end{document}